\documentclass[final]{cvpr}

\usepackage{times}
\usepackage{epsfig}
\usepackage{graphicx}
\usepackage{amsmath}
\usepackage{amssymb}

\usepackage[pagebackref=true,breaklinks=true,colorlinks,bookmarks=false]{hyperref}

\def\cvprPaperID{1234} 
\def\confYear{CVPR 2021}

\usepackage{amsmath}
\usepackage{amssymb}
\usepackage{pifont}
\usepackage{algorithm}  
\usepackage{algorithmic}  
\usepackage{color}
\usepackage{multirow}
\usepackage[all]{xy}
\usepackage{wrapfig}
\usepackage{picinpar}
\usepackage{booktabs} 
\newenvironment{Algorithm}[2][tbh]%
{
\centering
\begin{minipage}{#2}
\begin{algorithm}[H]}%
{\end{algorithm}
\end{minipage}
}

\begin{document}

\title{Few-shot Continual Learning: a Brain-inspired Approach}

\author{Liyuan Wang\textsuperscript{\rm 1 2 3}, Qian Li\textsuperscript{\rm 1 2}, Yi Zhong\textsuperscript{\rm 1 2 *} and Jun Zhu\textsuperscript{\rm 3 *}\\

\textsuperscript{\rm 1} School of Life Sciences, IDG/McGovern Institute for Brain Research, \\ Tsinghua University, Beijing, China. 
\textsuperscript{\rm 2} Tsinghua-Peking Center for Life Sciences, \\ Beijing, China.
\textsuperscript{\rm 3} Dept. of Comp. Sci. \& Tech., Institute for AI, BNRist Center, \\ 
THBI Lab, Tsinghua University, Beijing, China. 
\textsuperscript{\rm *} \textit{Corresponding authors}
}

\maketitle

\begin{abstract}
It is an important yet challenging setting to continually learn new tasks from a few examples. Although numerous efforts have been devoted to either continual learning or few-shot learning, little work has considered this new setting of few-shot continual learning (FSCL), which needs to minimize the catastrophic forgetting to the old tasks and gradually improve the ability of few-shot generalization. In this paper, we provide a first systematic study on FSCL and present an effective solution with deep neural networks. Our solution is based on the observation that continual learning of a task sequence inevitably interferes few-shot generalization, which makes it highly nontrivial to extend few-shot learning strategies to continual learning scenarios. We draw inspirations from the robust brain system and develop a method that (1) interdependently updates a pair of fast / slow weights for continual learning and few-shot learning to disentangle their divergent objectives, inspired by the biological model of meta-plasticity and fast / slow synapse; and (2) applies a brain-inspired two-step consolidation strategy to learn a task sequence without forgetting in the fast weights while improve generalization without overfitting in the slow weights. Extensive results on various benchmarks show that our method achieves a better performance than joint training of all the tasks ever seen. The ability of few-shot generalization is also substantially improved from incoming tasks and examples.
\end{abstract}


\section{Introduction}

Few-shot learning (FSL) and continual learning (CL) are two important yet challenging tasks, which usually co-exist in scientific and engineering domains. For instance, an AI-assisted diagnosis system needs to continually learn from a few cases to provide diagnostic suggestions and incrementally improve its ability of diagnosis. However, FSL and CL are basically studied as two separate directions. FSL aims to learn a novel task from a few examples. A general strategy of FSL is to learn transferable knowledge from large amounts of base tasks, and then generalize to a few examples of a novel task. Following this idea, \textit{gradient-based FSL} learns to initialize parameters for fast adaptation to a few examples \cite{finn2017model}; \textit{metric-based FSL} learns a generalized embedding space to match the representations of target examples to support examples by distance metrics \cite{snell2017prototypical,chen2019closer}. On the other hand, CL attempts to learn a sequence of novel tasks by avoiding \textit{catastrophic forgetting} \cite{mcclelland1995there}, that a neural network catastrophically forgets the learned parameters for old tasks when accommodating for new tasks. To mitigate catastrophic forgetting, effective strategies include regularization of important parameters for the old tasks, and replay of the old data distributions \cite{parisi2019continual}.

\newcommand{\tabincell}[2]{\begin{tabular}{@{}#1@{}}#2\end{tabular}}
\begin{table}[h]
  \vspace{-.1cm}
	\caption{Comparison of settings that consider both continual learning and few-shot learning.}\smallskip
    \vspace{-.2cm}
	\centering
	\resizebox{0.9\columnwidth}{!}{
		\smallskip
		\begin{tabular}{ccc}
		\specialrule{0.01em}{1.2pt}{1.5pt}
		  &\tabincell{c}{Objective of CL} &\tabincell{c}{Objective of FSL} \\
         \specialrule{0.01em}{1.2pt}{1.5pt}
         OML \cite{javed2019meta} &  \ding{51} & \ding{55} \\
         CBCL \cite{ayub2020brain} & \ding{51} & \ding{55} \\
         TOPIC \cite{tao2020few} & \ding{51} & \ding{55} \\
         AAN \cite{ren2018meta}   & \ding{55} & \ding{51}  \\
         IDA \cite{liu2020incremental}  & \ding{55} & \ding{51}  \\
         Ours  & \ding{51} & \ding{51}\\
		\specialrule{0.01em}{1.2pt}{1.5pt}
	\end{tabular}
}
	\label{Setting}
	\vspace{-.1cm}
\end{table}

A few very recent efforts start to consider the challenges of FSL and CL together. Although all of them try to continually learn from a few examples, they focus on either continually tackling a sequence of new tasks without forgetting (i.e. the objective of CL), or improving the ability to learn a novel task with a few examples (i.e. the objective of FSL), rather than achieving the two objectives together. Table \ref{Setting} summarizes the aims of current settings. 
Specifically, OML \cite{javed2019meta} learns representations that are robust to catastrophic forgetting, but only works well in relatively simple datasets with online training. CBCL \cite{ayub2020brain} stores the clustered feature embedding of incremental classes for joint prediction. TOPIC \cite{tao2020few} applies a neural gas network to learn first several tasks from large amounts of data, followed by several new tasks using a few examples. AAN \cite{ren2019incremental} uses an attention attractor network to learn a few new classes without interfering the performance on the large amounts of base classes. IDA \cite{liu2020incremental} improves FSL from incremental examples by indirect discriminant alignment.

\begin{figure}[ht]
	\centering
	\includegraphics[width=1\linewidth]{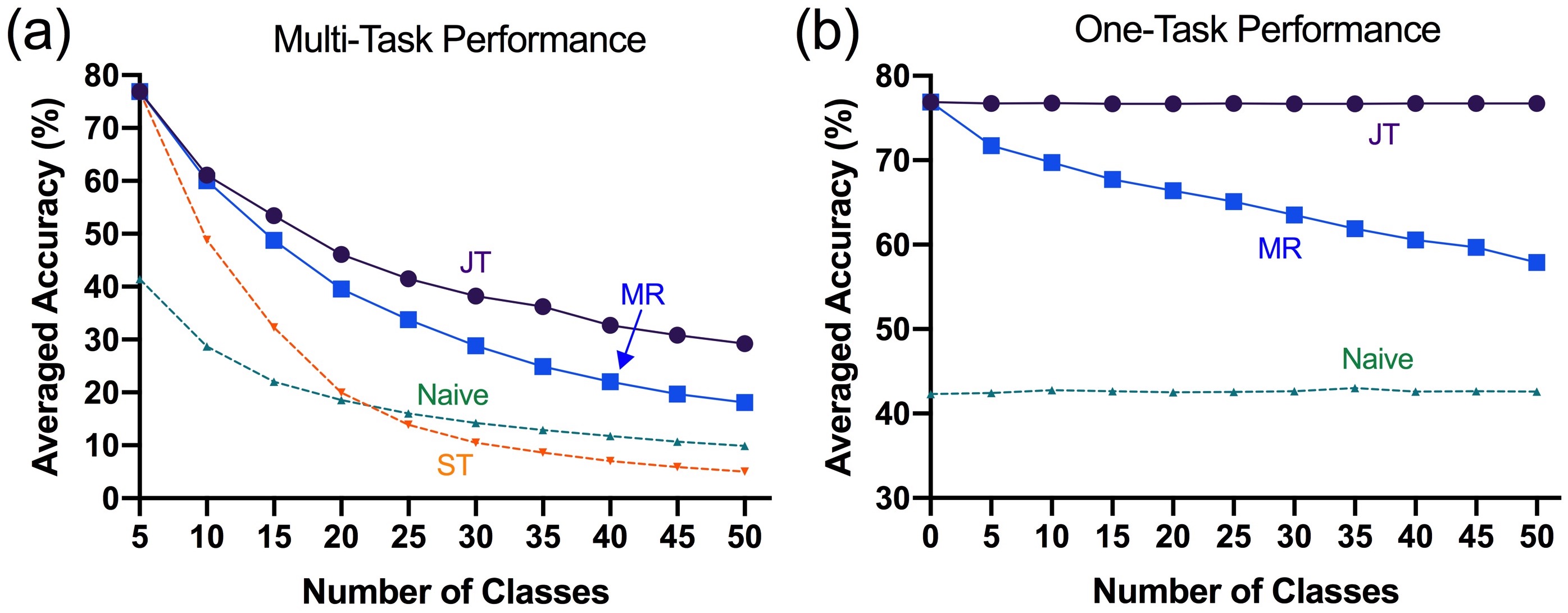}
	\vspace{-0.1cm}
	\caption{The ability of FSL is interfered by CL on miniImageNet. (a) is the performance to infer all the incremental classes ever seen. (b) is the performance on one 5-way 5-shot task during continual learning in (a). JT: Joint Training; ST: Sequential Training; MR: Memory replay of the training data ever seen; Naive: A randomly-initialized network.}
	\label{FSL_in_CL_Accuracy}
    \vspace{-0.1cm}
\end{figure}

Ideally, FSL and CL should benefit each other. The ability of FSL enables a model to learn incoming tasks from a few examples, and CL of new tasks and examples provides additional data sources to improve generalization. From analyzing the limitations of current efforts, we propose a more challenging but realistic setting as few-shot continual learning (FSCL): A model continually learns from a few examples and achieves the objectives of CL and FSL together, that: (1) tackle a sequence of novel tasks while avoiding catastrophic forgetting; and (2) improve FSL from incoming tasks and examples while avoiding overfitting. However, the proposed objectives of FSL and CL are not easy to achieve simultaneously, because CL tries to precisely memorize incoming tasks to mitigate forgetting, which inevitably ``overfits'' the task sequence and interferes FSL. If the model of FSCL is not properly designed, e.g., to simply combine the strategies of FSL and CL, the learned ability of FSL is not improved from CL but substantially decreased (Fig. \ref{FSL_in_CL_Accuracy}, b), and thus interferes the performance on the entire task sequence (Fig. \ref{FSL_in_CL_Accuracy}, a).

Compared with artificial neural networks, the biological brain is able to continually learn from everyday occurrences and incrementally improve generalization \cite{mcclelland2013incorporating, o2014selective, roxin2013efficient}. Catastrophic interference can be effectively mitigated by replaying old experience together with new inputs \cite{mcclelland1995there,mcclelland2013incorporating}. To fast generalize to an experience without overfitting, biological synapses interdependently update a fast and a slow components of plasticity, that a fast state quickly learns various inputs and then updates a slow state with a small stepsize \cite{roxin2013efficient,fusi2017computational}. Such synaptic plasticity across brain areas is modulated by replay activity \cite{fusi2017computational}. Particularly, a replay strategy named \textit{two-step consolidation} \cite{o2014selective} addresses the challenges of CL and FSL simultaneously: The first step is a strong input of the current occurrences to tag the synaptic activities (\textit{tagging}). While, the second step is to update the network with multiple weak replay inputs, where the cumulative activity of a synapse is applied to gate its stability (\textit{capture}) and thus to selectively generalize memories.

Inspired by the model of fast / slow synapse, we interdependently update two sets of parameters, i.e. fast weights and slow weights, to disentangle the divergent objectives of CL and FSL. Inspired by the model of two-step consolidation, we establish a replay system to learn the fast / slow weights: The first step is to approximate the magnitude of expected gradients for incoming tasks in the slow weights, as the ``activity tag''. The second step is to update the fast weights for multiple iterations, with regularization on the gated cumulative activity to balance fitting and generalization. Then, the slow weights are updated for a small step by the fast weights to improve generalization. We sample task sequences and training examples from various benchmark datasets of FSL, including Omniglot, miniImageNet, tieredImageNet and CUB-200-2011, as the evaluation benchmarks for FSCL. Extensive evaluations on such benchmarks support the effectiveness of our method to achieve the proposed aims of FSCL.

Our contributions include: 1. We propose a challenging but realistic setting as FSCL, in which the objectives of FSL and CL should be achieved simultaneously to improve each other, and provide a systematic analysis to disclose the relation between FSL and CL in FSCL; 2. We draw inspirations from the biological models of synaptic plasticity and memory replay to propose a novel approach for FSCL, named two-step consolidation (TSC), to interdependently update parameters for FSL and CL, respectively; 3. Extensive evaluations on benchmarks of FSCL validate the effectiveness of our strategy to mitigate catastrophic forgetting and incrementally improve FSL from CL.

\section{Problem Formulation}
We start by introducing the notations and setup of few-shot continual learning (FSCL). Then we analyze the nontrivial interference between CL and FSL in FSCL.

\subsection{Setup and Problem Formulation}

Few-shot continual learning (FSCL) is a setting where a learner continually learns \(T\) tasks and each task has a few training examples. For a typical classification task, let $N$ be the number of classes and $K$ be the number of training examples for each class.\footnote{It is straightforward to have different $K$ for various classes.} We call such a task $N$-way $K$-shot classification. Let \(D_{t} = \{(x_{i}^{t}, y_{i}^{t})\}_{i=1}^{N \times K} \) be the fully-labeled training set of task $t$. Then, the collection \( D_T = \bigcup_{t=1}^{T}D_{t}\) is called a query dataset for learning on the task sequence. In FSCL, \(D_T\) is sequentially provided to the learner, that \(D_{t}\) is introduced when training on task \(t\).

An ideal model for FSCL should achieve the objectives of both few-short learning (FSL) and continual learning (CL) simultaneously --- it can quickly learn a new task from a few examples, mitigate catastrophic forgetting of the old tasks, and incrementally improve FSL from incoming tasks and examples. Correspondingly, the evaluation should reflect the two objectives --- it should perform single-head evaluation \cite{chaudhry2018riemannian}, i.e. all the learned tasks are evaluated without access to the task label, to test the ability of CL; and also evaluate on a new task with a few training samples to test the ability of FSL. Table \ref{Setting} summarizes the difference of FSCL from the existing studies, which simply improve one of them.

\begin{figure*}[ht]
	\centering
    \vspace{-0.1cm}
	\includegraphics[width=1\linewidth]{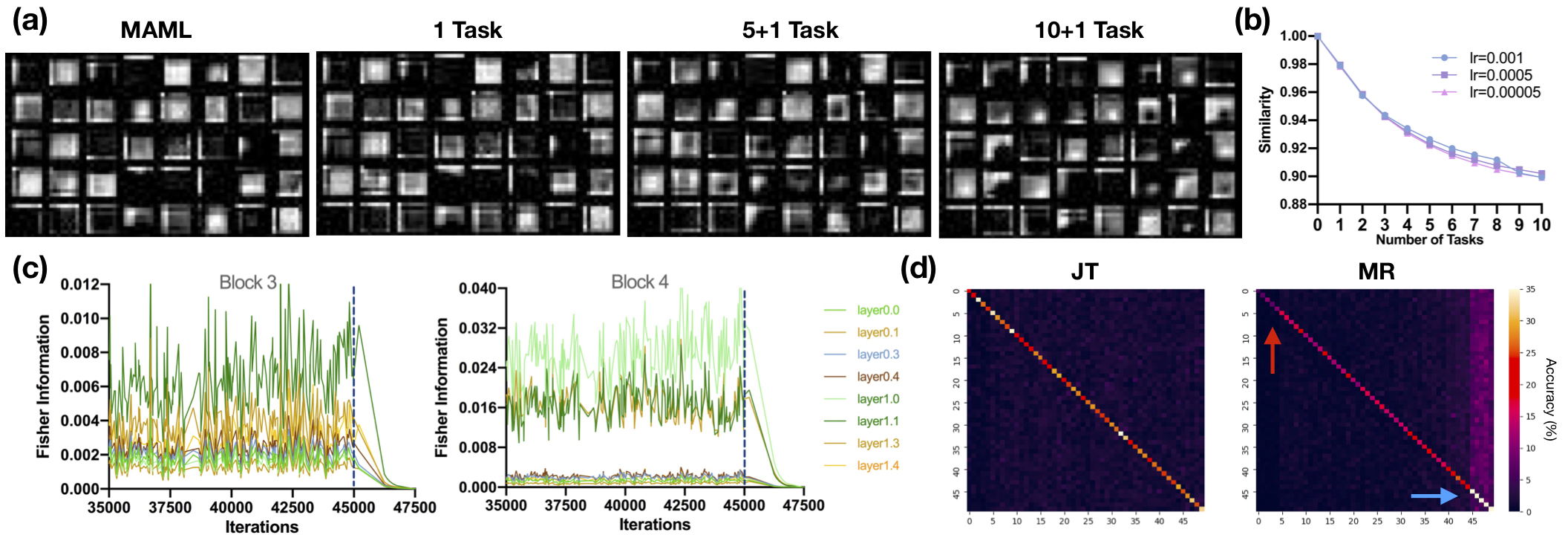}
	\caption{The objectives of FSL and CL interfere with each other in FSCL on miniImageNet. (a, b) The learned feature embedding for FSL is incrementally deformed in CL on various learning rates (lr).
(c) Fisher Information of feature extractor parameters on ResNet-18 in meta-training (before 45k iterations) and training on a specific task (after 45k iterations). (Best viewed in color) (d) Confusion matrix of CL of ten 5W5S tasks. The blue and the red arrows indicate overfitting to the current task and forgetting of the earlier tasks, respectively. x-axis: predicted labels; y-axis: ground-truth labels.}
	\label{FSL_in_CL_Visualization}
    \vspace{-0.1cm}
\end{figure*}

\subsection{Interference between Continual Learning and Few-Shot Generalization in FSCL}

We now analyze the nontrivial interference between continual learning and few-shot generalization if the FSCL model is not properly designed. 

To solve FSCL, a straightforward idea would be to directly combine the strategies of FSL and CL to build a model. Specifically, we extend the state-of-the-art FSL models to a continual learning setting. Here, we choose gradient-based FSL (e.g., MAML \cite{finn2017model}) as the base model for FSCL. \footnote{We consider metric-based FSL as the base model for FSCL in Appendix A.} The entire training process includes two stages: The model first learns transferable knowledge from large amounts of base tasks in a support set \(S\), called a meta-training (\textit{MT}) stage. \(S\) should be a much larger dataset with transferable domains to \(D_T\) while without overlapped classes and examples (\(S \bigcap D_T = \emptyset\)): e.g., if \(D_T\) is a dataset of birds, \(S\) can be a larger dataset of other animals.
Then, the model continually learns a sequence of novel tasks from \(D_T\) without access to \(S\), called a \textit{FSCL} stage. CL strategies, like memory replay and weight regularization, are implemented to mitigate catastrophic forgetting. After training of each task, the performances of both CL and FSL are evaluated. 

However, to precisely memorize the incoming tasks inevitably ``overfits'' the tasks ever seen in the FSCL stage, which interferes the ability of FSL on following tasks. Fig. \ref{FSL_in_CL_Accuracy} shows the results of FSCL on miniImageNet. Joint training (JT) of all the tasks ever seen is the upper-bound performance for the objective of CL. Compared with sequential training (ST) of each task, memory replay (MR) of the training data ever seen can completely avoid catastrophic forgetting, but significantly underperforms JT (Fig. \ref{FSL_in_CL_Accuracy}, a). The ability of FSL also gradually decreases from learning the task sequence (Fig. \ref{FSL_in_CL_Accuracy}, b). 
Either, a simple combination of regularization-based CL methods cannot solve this issue (Table \ref{FSCL_Overall_Accuracy}).

To explicitly show the interference of CL to FSL, we first visualize the feature embedding in CL of several few-shot classification tasks on miniImageNet with MR to avoid catastrophic forgetting. As shown in Fig. \ref{FSL_in_CL_Visualization} (a, b), the learned feature embedding is only slightly adjusted to learn a new 5-way 5-shot (5W5S) task. However, after learning several additional tasks and then training on the 5W5S task, the learned feature embedding is incrementally deformed, which interferes the performance of FSL. Next, when the model moves from the MT stage to the FSCL stage, we observe a significant collapse of Fisher Information of the feature extractor parameters (Fig. \ref{FSL_in_CL_Visualization}, c), which indicates forgetting of the important parameters for FSL \cite{kirkpatrick2017overcoming}. Further, the confusion matrix of the above experiments indicates that CL of new task interferes generalization to the entire task sequence (Fig. \ref{FSL_in_CL_Visualization}, d), which cannot be addressed by a simple combination of MR and weight regularization (Fig. \ref{FSCL_confusion}).

\begin{figure}[ht]
	\centering
	\includegraphics[width=1\linewidth]{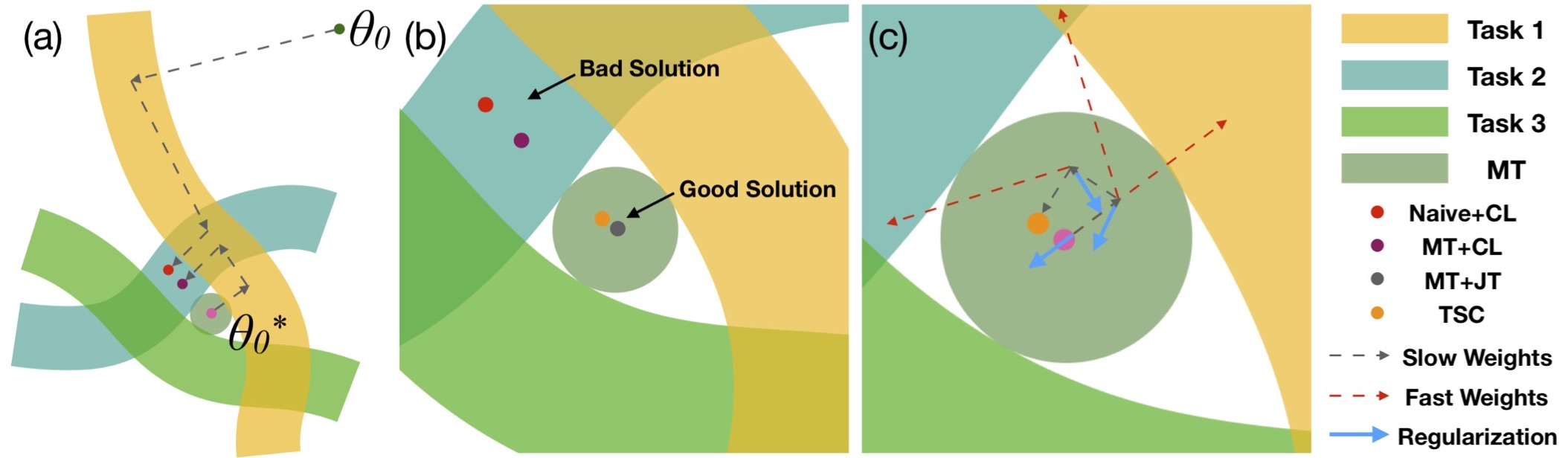}
    \vspace{-0.1cm}
	\caption{A more formal illustration of FSCL and our proposal. (a) \(\theta_{0}^{*}\) and \(\theta_{0}\) are the parameters of a FSL base model and a naive model, respectively. In continual learning, they finally converge to the bad solution manifold in (b). (c) is our solution to adjust the parameters to the good solution manifold shown in (b). MT: Manifold learned from the MT stage; JT: Joint Training; CL: Continual Learning.}
	\label{FSCL_Theoretical_Model}
    \vspace{-0.1cm}
\end{figure}

We provides a slightly more formal illustration of FSCL in Fig. \ref{FSCL_Theoretical_Model} (a, b), where \(\theta_{0}^{*}\) and \(\theta_{0}\) are the parameters of a network with or without training on the support set \(S\), respectively. After training on large amounts of base tasks sampled from \(S \), \(\theta_{0}^{*}\) moves to the center of the solution manifolds of these tasks, and thus can quickly generalize to a novel task from a few examples. We assume a CL approach can completely avoid catastrophic forgetting. However, after continual learning of several tasks, \(\theta_{0}^{*}\) finally converges to a solution manifold close to \(\theta_{0}\) with poor performance, i.e. bad solution. By contrast, if the distributions of \(S\) and \(D_T\) are the same and all \(T\) tasks are \textit{i.i.d.}, good solution of FSCL should be close to the center. If the distributions of \(S\) and \(D_T\) are (slightly) different, the optimal solution will (slightly) deviate the center. Thus, the network should be carefully adjusted on the sequentially arrived \(D_t\) to improve generalization while avoiding overfitting. 

\section{Our Proposal}
We now present our solution to address the above issues that draws inspirations from biological brain --- an ideal system for FSCL compared with artificial neural networks.

\subsection{Two-Step Consolidation}

\begin{figure}[ht]
	\centering
	\includegraphics[width=1\linewidth]{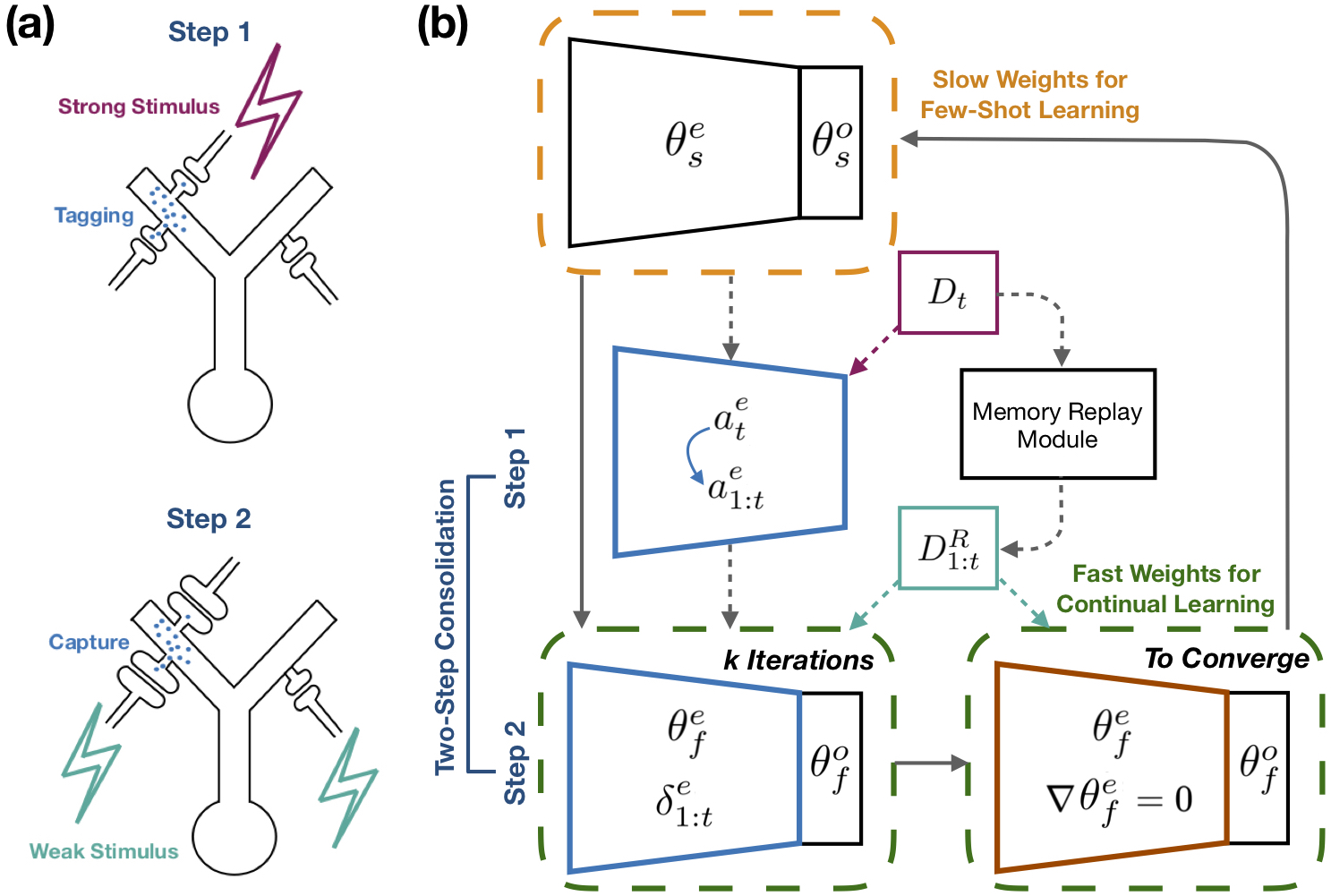}
	\caption{The biological and computational models of  two-step consolidation (TSC). (a) Synapses are tagged by a strong input in the first step and then captured by multiple weak inputs in the second step. (b) Architecture of TSC to interdependently update the fast / slow weights for CL and FSL, respectively.} 
	\label{TSC}
   \vspace{-0.1cm}
\end{figure}

Biological brain can continually learn new tasks from a few examples and apply the learned information for generalization \cite{o2014selective,mcclelland2013incorporating,lewis2011overlapping}. Replay of old memory with new inputs is able to mitigate catastrophic forgetting \cite{mcclelland2013incorporating}.
Particularly, a replay strategy named \textit{two-step consolidation} (TSC) has been shown effective to integrate a few examples to a network and selectively generalize memories \cite{o2014selective} (Fig. \ref{TSC}, a). 
The first step is a strong input of the current occurrence to tag the synaptic activity. A \textit{synaptic tagging and capture} (STC) process selectively gates the stability of synapses dependent on the cumulative tags. Then, multiple weak replay inputs in the second step can selectively generalize the learned memories for inference \cite{o2014selective}. Correspondingly, the biological synaptic strength is modulated by replay activity through two components of plasticity, which depend on the history of synaptic modifications (called \textit{meta-plasticity}): The state of synaptic strengths is copied from the previous state. A fast component quickly encodes various replay inputs, and then updates a slow component by a small step as the current state, which will be copied to the downstream states \cite{fusi2017computational}. 

\subsection{Our Solution to FSCL}
Inspired by such brain strategies to continually generalize memories to incoming occurrences and learn from them, we propose a novel solution to FSCL named TSC, which enables a network to simultaneously achieve the objectives of FSL and CL (Fig. \ref{TSC}, b and Algorithm 1). To disentangle the divergent objectives of FSL and CL, we use two sets of parameters for a network, i.e. {\it fast weights} \(\theta_{f}\) and {\it slow weights} \(\theta_{s}\), and interdependently update them for CL and FSL, respectively. We consider a typical network for FSCL that includes a feature embedding layer (\(e\)) and an output layer (\(o\)). Correspondingly, both \(\theta_{f}\) and \(\theta_{s}\) consist of the parameters of \(e\) and \(o\) (if applicable) that \(\theta_f = \theta_f^{e} \cup \theta_f^{o}\), \(\theta_s = \theta_s^{e} \cup \theta_s^{o}\). For gradient-based few-shot classification, the output layer is a linear classifier (or fully connected layer) in general. Our method draws key inspirations from how biological synapses are modulated by replay inputs to precisely memorize the old experience and better generalize to a new experience. As shown in Fig. \ref{TSC} (b), we first implement a memory replay module, and then apply a brain-inspired two-step consolidation (TSC) strategy to interdependently update \(\theta_f\) and \(\theta_s\) on a replay dataset, as detailed below.

\begin{Algorithm}[ht]{8.3cm}
	\caption{Algorithm of TSC for FSCL}
	\begin{algorithmic}[1]
        \STATE \textbf{Require:} \(\theta_{s}\): parameters learned from the MT stage; \(D_{t}\): training dataset of task \(t\); \(\beta\), \(\lambda\), \(m\): hyperparameters; \(k\): iterations to learn \(\theta_f^{e}\).
        \FOR{task \(t = 1, 2, 3, ...\)} 
		\STATE Establish the replay dataset \(D_{1:t}^{R} = D_{t} \bigcup\ {D}_{\le t-1}^{R}\)
        \STATE Step 1: Compute \(a_{t}^{e}\) and update \(a_{1:t}^{e}\), \({\delta}_{1:t}^{e}\) 
        \STATE \(\theta_{f}\)  \(\leftarrow\) \(\theta_{s}\) 
        \FOR{\(1, 2, ..., k\)} 
        \STATE Step 2: Update \(L_{e}\,(\theta_f^{e}, \theta_f^{o}, D_{1:t}^{R})\)
        \ENDFOR
        \WHILE{not done}
        \STATE Update \(L_{o}\,(\theta_f^{o}, D_{1:t}^{R})\) 
        \ENDWHILE
        \STATE \(\theta_{s}\)  \(\leftarrow\) \(\theta_{s}\)  - \(\beta \, (\theta_{s} - \theta_{f}) \)
        \ENDFOR
	\end{algorithmic}
\end{Algorithm}
\vspace{+0.1cm}

To mitigate catastrophic forgetting of the old experience, we apply a memory replay module to provides a replay dataset of the old tasks as \({D}_{\le t-1}^{R}\) when learning a new task \(t\). This module can be a memory buffer to replay old training data, or a generative model to learn the old data distributions and replay generated data. 

To effectively learn the parameters, we need to consider the trade-off between fitting and generalization. We first propose a strategy to interdependently update \(\theta_{f}\) and \(\theta_{s}\), inspired by the biological fast / slow synapse. In the MT stage, \(\theta_{s}\) are first trained on a support set \(S\). In the FSCL stage, \(\theta_{s}\) transfer parameters to \(\theta_{f}\) before training on each task. \(\theta_{f}\) learn incoming tasks and examples for inference on the task sequence. \(\theta_{s}\) are updated by \(\theta_{f}\) for a small step \(\beta\) to improve the ability of generalization from the new tasks and examples. \(\theta_{s}\) then transfer the updated parameters to \(\theta_{f}\) to learn the next task. In such a framework, \(\theta_{f}\) tend to precisely learn and memorize the task sequence for better inference, but should only feedback generalized information to \(\theta_{s}\). 

To achieve this aim, we learn \(\theta_f^{e}\) on \(D_{1:t}^{R}=D_{t} \bigcup\ {D}_{\le t-1}^{R}\) for limited \(k\) iterations and regularize the training: large changes of the parameters should be penalized to avoid overfitting, while a too strong limitation generally results in underfitting.
We draw inspirations from the biological {synaptic tagging and capture} (STC) in two-step consolidation: Synaptic activities in a network are first tagged by a new input of the current experience. Then the network is updated by multiple replay inputs, where the synaptic stability is gated by the cumulative activity tags (Fig. \ref{TSC}, a). Specifically, we propose to learn \(\theta_{f}^{e}\) with a STC regularization strategy, which penalizes changes of \(\theta_{f}^{e}\) from \(\theta_s^{e}\) dependent on the gated cumulative activity of learning history:

\begin{equation}
\begin{split}
L_{e}(\theta_f^{e}, \theta_f^{o}, D_{1:t}^{R}) = 
 L_{(x, y)\sim D_{1:t}^{R} }\, (\theta_f^{e}, \theta_f^{o}) \\ 
 +\lambda \, \sum_l \sum_{i=1}^{N_l}{\delta}_{1:t, i}^{e, l}\,( \theta_{f, i}^{e, l} - \theta_{s, i}^{e, l})^{2},
\end{split}
\end{equation}

\begin{equation}
 {\delta}_{1:t, i}^{e, l} = \sigma [m ({a}_{1:t, i}^{e, l}-\frac{1}{t}\frac{1}{N_l} \sum_{i=1}^{N_l} {a}_{1:t, i}^{e, l})],
\end{equation}
where \(N_l\) is the number of parameter \(i\) in layer \(l\), \({a}_{1:t, i}^{e, l}\) is the cumulative activity in task \(1:t\) of the parameter, and \(m\) is a positive scale factor. For task \(t\), we calculate the magnitude of the expected gradients as the ``activity'' of a parameter in \(\theta_s^{e}\), that \(a_{t, i}^{e, l} = \parallel\mathbb{E}_{(x, \, y)\sim {D}_{t}}\,[ \nabla{L}(\theta_{s, i}^{e, l}) ] \parallel\). The cumulative activity is updated as \({a}_{1:t, i}^{e, l} = {a}_{1:t-1, i}^{e, l}+  {a}_{t, i}^{e, l}\). For each parameter \(i\) in layer \(l\), we average \({a}_{1:t, i}^{e, l}\) in that layer of \(t\) tasks as the ``threshold'', and use a sigmoid function \(\sigma\) as the ``gate'' to project the threshold-centered cumulative activity to the values in \((0,1)\). 
Therefore, the changes of \(\theta_{f}^{e}\) are limited by their cumulative magnitude of the expected gradients to alleviate overfitting, while the sigmoid function clips the penalty to avoid underfitting. \(L_{(x, y)\sim D_{1:t}^R }\)  is the loss function for the target tasks trained on the replay dataset \(D_{1:t}^{R}\). We use a cross entropy loss for classification. 

Then we update \(\theta_{f}^{o}\) with fixed \(\theta_{f}^{e}\) for better inference on the task sequence, which can be multiple iterations till convergence:
\begin{equation}                                           
L_{o}(\theta_f^{o}, D_{1:t}^{R}) =  L_{(x, y)\sim D_{1:t}^{R} }\, (\theta_f^{o}).
\end{equation}
The overall objective of our method is defined as: 
\begin{equation}\label{eq:TSC}
\begin{split}
L_{e}(\theta_f^{e}, \theta_f^{o}, D_{1:t}^{R}) 
 + L_{o}(\theta_f^{o}, D_{1:t}^{R}), ~ t = 1, 2, ... , T.
\end{split}
\end{equation}
As shown in Fig. \ref{FSCL_Theoretical_Model} (c), the brain-inspired designs in TSC provide an effective solution to FSCL:
The fast / slow weight strategy enables a network to fit a task sequence by fast weights, but only update slow weights by a small step to avoid overfitting. Then, the regularization strategy further regularizes the changes of parameters to the direction of good solution manifold.


\section{Experiment}
\subsection{Experiment Setup}
Most methods of gradient-based FSL, e.g., MAML \cite{finn2017model}, MAML++\cite{antoniou2018train}, with linear classifier \cite{chen2019closer} can be easily adapted to TSC. 
We also consider metric-based FSL as the base model by implementing ProtoNet \cite{snell2017prototypical} in Appendix A. 
We evaluate TSC for FSCL of \textit{New Class} and \textit{New Instance} \cite{lomonaco2017core50,antoniou2020defining}, first on three benchmark datasets of FSL, i.e. Omniglot, miniImageNet and tieredImageNet. Then we consider huge domain differences, i.e. \( S\) is the training set of tieredImageNet while \(D_T\) is sampled from CUB-200-2011 dataset (referred to as the CUB). We apply the averaged accuracy (\({A}_{t}\)) on the test set of \(t\) tasks trained so far to evaluate the ability of CL, and evaluate the ability of FSL on an additional novel task. All the experiment results are averaged by 20 randomly sampled tasks or task sequences, if not specified.

\textbf{Dataset:} Omniglot \cite{lake2011one} is a dataset of handwritten characters, consisting of 963 classes for training and 660 classes for testing with 20 images per class . miniImageNet \cite{vinyals2016matching} is a dataset of color images derived from the 1000-class ILSVRC-12 dataset \cite{russakovsky2015imagenet}, including 100 classes with 600 images per class. We follow the same split of miniImageNet from ILSVRC-12 as \cite{vinyals2016matching}, but apply all 100 classes for FSCL and the remained 900 classes as the support set. tieredImageNet \cite{ren2018meta} is also derived from  ILSVRC-12, including 608 classes from 34 super-classes. Each super-class consists of 20, 6, and 8 classes for training, validation and testing, respectively. CUB-200-2011 \cite{wah2011caltech} is a dataset including 200 classes and 11,788 colored images of birds. Omniglot and other three datasets are resized to \(32 \times 32\) and \(64 \times 64\) respectively before experiments \cite{antoniou2020defining}. Since the hierarchy in tieredImageNet splits the testing classes sufficiently distinct from the training classes \cite{ren2018meta}, domain differences between \(S\) and \(D_T\) are generally larger and larger from miniImageNet, tieredImageNet to tieredImageNet\(\rightarrow\)CUB.

\begin{table*}[ht]
	\centering
	\caption{Averaged accuracy (\%) on a sequence of 5W5S classification tasks. All the baselines and TSC are initialized by MAML \cite{finn2017model}. The results are shown as top-1 accuracy or top-1 / top-5 accuracy.}\smallskip
	\resizebox{1\textwidth}{!}{
	\begin{tabular}{clcccccccc}
		\specialrule{0.01em}{1.2pt}{1.5pt}
		\multicolumn{2}{c}{} & \multicolumn{2}{c}{Omniglot} & \multicolumn{2}{c}{miniImageNet} &
		\multicolumn{2}{c}{tieredImageNet} & \multicolumn{2}{c}{tieredImageNet\(\rightarrow\)CUB}\\
		&Method & $ A_{30} $ & $ A_{50} $  & $ A_{30} $  & $ A_{50} $ & $ A_{30} $ & $ A_{50} $& $ A_{30} $ & $ A_{50} $\\
        \specialrule{0.01em}{1.2pt}{1.5pt}
		\multirow{6}*{\tabincell{c}{Baseline}}
	    &JT& 96.20&94.97&36.83 / 70.46&28.75 / 59.60 &31.51 / 65.34&23.04 / 53.76 & 29.83 / 64.58&23.43 / 53.86\\
        &MR &94.38 &92.13& 28.85 / 54.39&18.07 / 38.42 &24.21 / 50.57&15.10 / 34.80&20.86 / 45.79&13.17 / 31.87 \\
        &GR & 94.93&93.49&28.49 / 55.02&19.81 / 41.45 &24.72 / 50.40&16.30 / 37.42&21.26 / 48.14&15.68 / 36.90\\
       &SI-M \cite{snell2017prototypical} &94.97 &94.51 &28.08 / 54.13&19.09 / 39.67&24.66 / 51.60&16.57 / 36.39&20.11 / 45.92&14.64 / 33.71\\
        &EWC-M \cite{kirkpatrick2017overcoming} & 95.52 & 93.95 &27.46 / 54.88&25.29 / 53.19 & 25.76 / 53.44 & 21.99 / 50.01 &23.97 / 57.47&15.87 / 41.97\\
        &MAS-M \cite{aljundi2018memory} & 95.81 & 94.47 & 28.94 / 60.48&27.01 / 58.98& 26.48 / 58.25 & 22.11 / 53.84 &22.33 / 54.41&18.60 / 47.84\\
        \specialrule{0.01em}{1.2pt}{1.5pt}
		\multirow{4}*{\tabincell{c}{TSC\\(Ours)}}
          &Reg & 95.84 & 95.06  &35.02 / 72.02&29.56 / 60.44 & 30.35 / 67.95 & 22.93 / 55.72 &27.24 / 61.48&20.48 / 48.82\\
         &FSW &96.44 &95.57 & 37.13 / 70.51&30.32 / 61.51 & 30.66 / 64.95&23.29 / 54.27 &29.75 / 63.89&23.76 / 54.29\\
        &Reg+FSW & \textbf{96.67} & \textbf{96.04} &\textbf{38.49} / \textbf{71.87}&\textbf{30.70} / \textbf{62.38}&\textbf{32.73} / \textbf{67.61}&\textbf{24.92} / \textbf{56.73}&\textbf{30.52} / \textbf{65.00}&\textbf{24.38} / \textbf{55.20}\\
		&Reg+FSW+GR &96.89 &96.63 & 39.23 / 72.56&33.66 / 61.55 & 33.33 / 67.99 & 25.48 / 56.43 &31.63 / 64.74&24.86 / 54.63 \\
		\specialrule{0.01em}{1.2pt}{1.5pt}
	\end{tabular}
	}
    \vspace{-.1cm}
	\label{FSCL_Overall_Accuracy}
\end{table*}

\begin{table}[ht]
	\centering
	\caption{Averaged accuracy (\%) of FSCL on a sequence of 5-way tasks on Omniglot. *APL uses 20-shot per class while other methods use 5-shot per class.}\smallskip
	\resizebox{0.80\linewidth}{!}{
	\begin{tabular}{lcc}
		\specialrule{0.01em}{1.2pt}{1.5pt}
		 Method & $ A_{30} $ & $ A_{50} $ \\
        \specialrule{0.01em}{1.2pt}{1.5pt}
          OML \cite{javed2019meta}& 91.47  & 88.55\\
          ANML \cite{beaulieu2020learning}& 96.71 & 95.47\\
          APL \cite{ramalho2019adaptive} & 94.88*  & 93.42*\\
        \specialrule{0.01em}{1.2pt}{1.5pt}
        MAML\cite{finn2017model}+JT& 96.20&94.97 \\
        MAML\cite{finn2017model}+MR &94.38 &92.13 \\
        MAML\cite{finn2017model}+TSC & 96.67 & 96.04 \\
        \specialrule{0.01em}{1.2pt}{1.5pt}
        MAML++\cite{antoniou2018train}+JT& 97.33&95.67 \\
        MAML++\cite{antoniou2018train}+MR &95.53 &93.26 \\
        MAML++\cite{antoniou2018train}+TSC &97.78 & 96.63 \\
        \specialrule{0.01em}{1.2pt}{1.5pt}
	\end{tabular}
	}
	\label{FSCL_Base_Model_Omn}
\end{table}

\textbf{Baseline:} 
First, we compare TSC with several baselines of representative CL methods, using MAML as the base model of FSL. Since replay-based methods show great advantages in single-head evaluation on incremental classes \cite{kemker2018measuring}, we consider the baselines as below:
(1) Joint training (JT) of all the tasks ever seen, i.e. the upper bound performance of CL.
(2) Memory replay (MR): We store the old training data ever seen in a memory buffer to theoretically avoid forgetting of the learned tasks. Although many existing work \cite{rebuffi2017icarl,castro2018end} use a fixed memory buffer (with a pre-reserved size of 20 images per class in general) selected from much larger amounts of training data, in FSCL the training data (typically 1- or 5-shot per class) is insufficient to perform selection. Also, the memory buffer of all the old data is much smaller than a fixed size of 20 images per class and is indeed competitive to commonly-used methods of selection as analyzed in~\cite{chaudhry2018riemannian}.
(3) MR with regularization-based methods: We combine MR with representative regularization-based methods (appended with '-M'), e.g., SI \cite{zenke2017continual}, EWC \cite{kirkpatrick2017overcoming} and MAS \cite{aljundi2018memory}, which penalize changes of the important parameters for the old tasks. 
(4) Generative replay (GR): To evaluate the strategy of generative replay and augment the limited training data, we train a generative model to recover the old data distributions.  

Next, we compare with several recent meta-learning approaches that improve the ability of CL through a better meta-training, including OML {javed2019meta}, ANML \cite{beaulieu2020learning} and APL \cite{ramalho2019adaptive}. OML \cite{javed2019meta} is a meta-learning approach to learn sparse representations that are more robust to catastrophic forgetting. ANML \cite{beaulieu2020learning} meta-learns an activation-gating function for context-dependent selective activation to improve OML. APL \cite{ramalho2019adaptive} learns a memory module from the support set for FSL, which also benefits CL since the memory module can be used for memory replay. In contrast to the methods above, TSC aims to extend a base model that achieves considerable ability of FSL to FSCL. We implement MAML++ \cite{antoniou2018train}, a more recent meta-learning approach for FSL, as the base model of TSC to show this advantage. For all experiments, we extensively evaluate hyperparameters and report the best performance.

\textbf{Architecture and Training:} The base model of MAML applies a ResNet-18 feature extractor with similar implementations of MAML as \cite{chen2019closer}, validated in Appendix B. For generative replay, we apply a 4-layer DCGAN architecture \cite{radford2015unsupervised} with WGAN gradient penalty to stabilize generation \cite{gulrajani2017improved}, but replace BatchNorm layers by InstanceNorm to learn a few examples. The implementations and hyperparameters are detailed in Appendix C.

\subsection{FSCL of New Class}
In Table \ref{FSCL_Overall_Accuracy}, we evaluate TSC on a sequence of 5-way 5-shot (5W5S) classification tasks. Because of the interference between CL and FSL, the performances of MR and GR are much lower than JT, although GR slightly outperforms MR due to the augmentation effects of generated data. The regularization-based methods penalize changes of the parameters and thus outperform MR, but still significantly underperform JT. In contrast, through interdependently updating the fast / slow weights (FSW) with the STC regularization (Reg), TSC substantially outperforms joint training (JT) and other baselines. The performance of TSC can be further improved by using GR as the memory replay module (Table \ref{FSCL_Overall_Accuracy}).

Then, we compare with the meta-learning approaches that improve the ability of CL through a better meta-training. As shown in Table \ref{FSCL_Base_Model_Omn}, although OML {javed2019meta} and ANML \cite{beaulieu2020learning} learn a sparse representation in meta-training to alleviate catastrophic forgetting in CL, the learned representation sacrifices the ability of FSL and thus decreases the performance on the entire task sequence. APL \cite{ramalho2019adaptive} requires a larger number of training data, e.g., 20-shot per class, to select prototypes in the memory modules, and significantly underperforms other methods. By contrast, TSC extends a base model that acquires a promising ability of FSL to FSCL. Therefore, TSC with MAML \cite{finn2017model} as the base model substantially outperforms other methods. The performance of FSCL can be further improved by using more recent meta-learning approaches for FSL as the base model, e.g., MAML++ \cite{antoniou2018train}.

\subsection{Ablation Study and Analysis}

We provide the ablation study of TSC in Table \ref{FSCL_Overall_Accuracy} that both the brain-inspired designs of the STC regularization (Reg) and the fast / slow weight (FSW) significantly improve FSCL. Here we provide a more extensive analysis of how the two strategies balance fitting and generalization in FSCL. First, we examine two hyperparameters to control FSW in Table \ref{Ablation_TSC}, including learning rate of the slow weight \(\theta_{s}\) (\(\beta\)) and iterations to learn the fast weight \(\theta_{f}\) (\(k\)). \(\beta = 0\) is equal to JT, while \(\beta = 1\) is a batch version of MR. Smaller \(\beta\) mitigates the decreased performance of CL through more slowly updating \(\theta_{s}\), while the performances on \(\beta = 0.01\) and \(\beta = 0.001\) are comparable. Then we evaluate \(k\) on \(\beta = 0.01\). Under the extreme condition of \(k=0\), i.e. the feature embedding layer is fixed, the model still achieves considerable performance and significantly outperforms MR (Table \ref{FSCL_Overall_Accuracy}), consistent with our analysis that the good solution of FSCL is located in the MT manifold if the generalization error is trivial. The network can be fully updated by a larger \(k\), but generally suffers from more overfitting. Compared with \(k=0\) and \(k=500\), the performance is improved by updating the feature extractor with limited iterations, i.e. \( k = 100\).

\begin{figure}[t]
	\centering
	\includegraphics[width=1\linewidth]{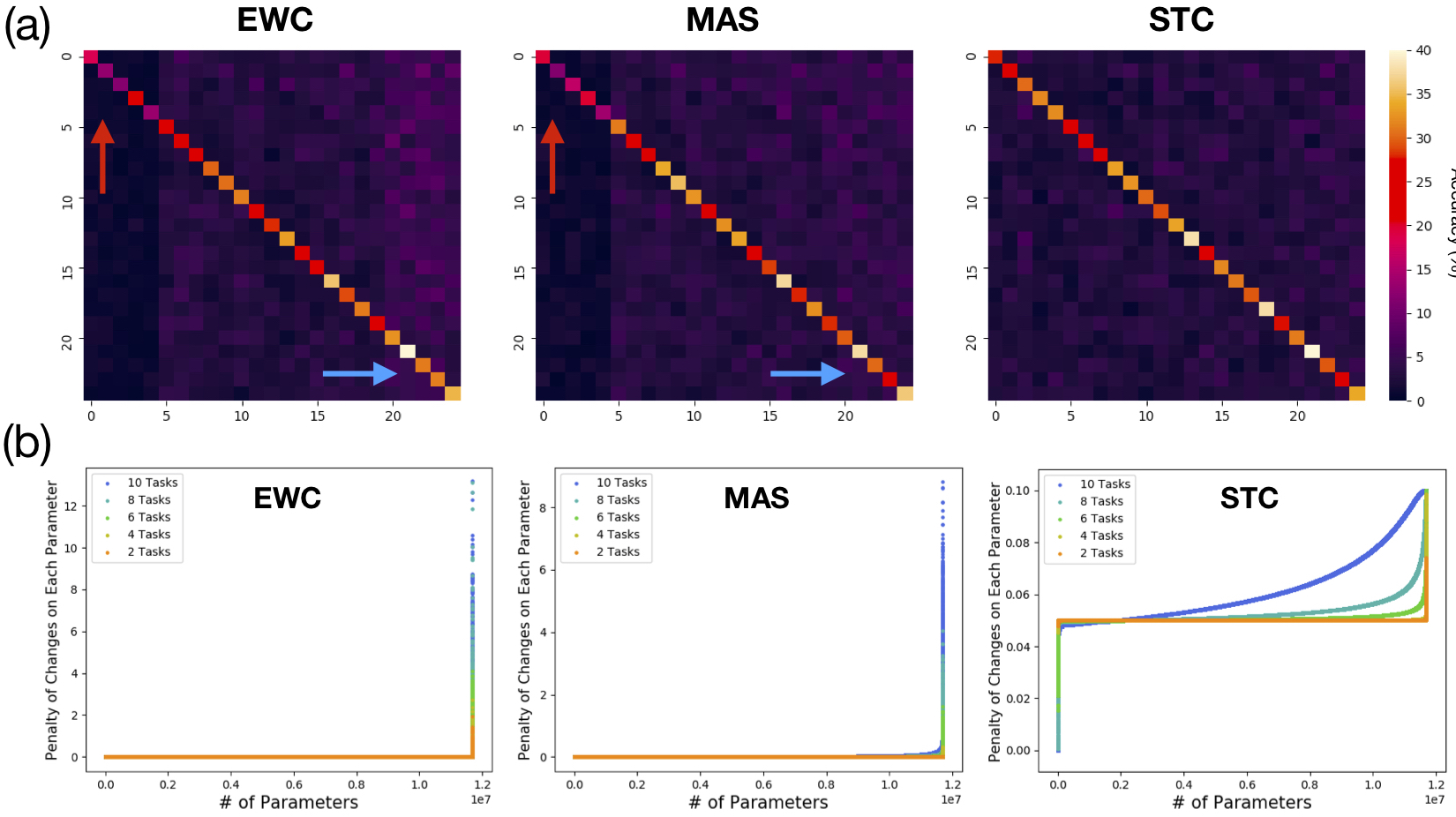}
	\caption{(a) Confusion matrix of various weight regularizations in FSCL of five 5W5S classification tasks on tieredImageNet. The blue and the red arrows indicate overfitting to the current task and forgetting of the earlier tasks, respectively. x-axis: predicted labels; y-axis: ground-truth labels. The distributions of penalty on the feature extractor parameters are shown in (b) (The scales of y-axis are different).}
	\label{FSCL_confusion}
   \vspace{-0.1cm}
\end{figure}

The STC regularization further addresses this issue by selectively stabilizing parameters, that \(k=100\)+Reg significantly outperforms JT, \(k=100\) and \(k=500\) (Table \ref{Ablation_TSC}). While adding a constant penalty (CP) cannot improve \(k=100\). On the other hand, the strategy of FSW with Reg further improves FSCL, compared with learning one-set of parameters in Table \ref{FSCL_Overall_Accuracy}. Next, we analyze how the STC regularization is more effective in FSCL than other regularization-based methods for CL. As shown in \ref{FSCL_confusion}, STC incrementally stabilizes parameters with a relatively small but constant penalty to balance fitting and generalization. While, other methods estimate a strong penalty on very few parameters, which cannot fit the entire task sequence well without interfering generalization (Fig. \ref{FSCL_confusion}, a, b). 

\begin{table}[th]
	\caption{Ablation study of TSC. We report averaged accuracy (\%) (top-1 / top-5) of 5W5S class-incremental learning. JT: joint training; MR: memory replay; Reg: the STC regularizer; CP: a constant penalty.}\smallskip
	\centering
	\resizebox{1 \columnwidth}{!}{
		\smallskip
		\begin{tabular}{clcccc}
		\specialrule{0.01em}{1.2pt}{1.5pt}
           & & \multicolumn{2}{c}{miniImageNet} & \multicolumn{2}{c}{tieredImageNet} \\
		  &Hyperparameters & $ A_{30} $  & $ A_{50} $ &  $ A_{30} $  & $ A_{50} $ \\
         \specialrule{0.01em}{1.2pt}{1.5pt}
          \multirow{5}*{\tabincell{c}{\(k=500\)}}
        &\(\beta = 0\) (JT) &36.83 / 70.46&28.75 / 59.60&31.51 / 65.34&23.04 / 53.76\\
         &\(\beta = 1\) (MR) &28.54 / 54.85 &18.17 / 38.58 & 24.65 / 50.73 &15.28 / 35.55\\
        & \(\beta = 0.1\) &35.25 / 66.64&24.17 / 50.96 & 28.75 / 61.03&19.28 / 44.47\\
         &\(\beta = 0.01\) &36.17 / 69.57&28.51 / 59.13 & 30.98 / 65.07&22.78 / 52.72\\
        & \(\beta = 0.001\) &35.85 / 69.17&28.52 / 59.69 & 31.17 / 65.41&22.84 / 52.70\\
         \specialrule{0.01em}{1.5pt}{1.5pt}
        \multirow{4}*{\tabincell{c}{\(\beta=0.01\)}}
         &\(k = 0\)  & 33.97 / 72.00&26.77 / 60.73 & 29.56 / 64.29&21.91 / 52.95\\
         &\(k = 100\) &37.13 / 70.51&30.32 / 61.51 & 30.66 / 64.95&23.29 / 54.27\\
        &\(k = 100\) +CP & 37.06 / 70.93&30.11 / 60.89&30.93 / 65.56&23.30 / 54.11\\
        &\(k = 100\) +Reg &\textbf{38.49} / \textbf{71.87}&\textbf{30.70} / \textbf{62.38}&\textbf{32.73} / \textbf{67.61}&\textbf{24.92} / \textbf{56.73}\\
		\specialrule{0.01em}{1.5pt}{1.5pt}
	\end{tabular}
}
	\label{Ablation_TSC}
	\vspace{-.1cm}
\end{table}

\begin{figure}[t]
	\centering
	\includegraphics[width=1\linewidth]{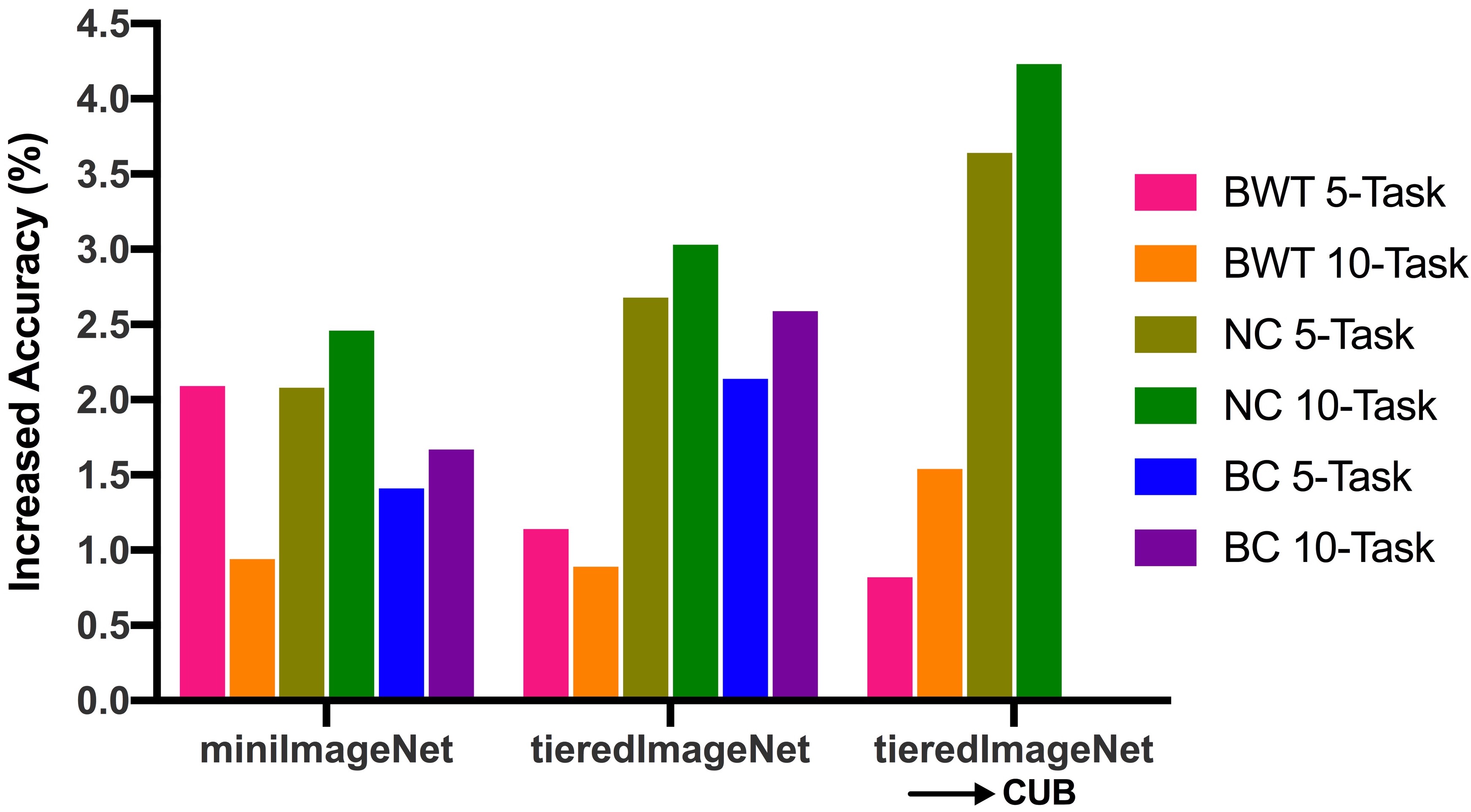}
	\caption{TSC improves the backward transfer (BWT, \%) and the 5W5S performance on both novel class (NC, \%) and base class (BC, \%) in FSCL, averaged by more than 20 randomly sampled task sequences.}
	\label{FSCL_Improve_FSL}
\end{figure}

\subsection{FSL is Incrementally Improved from CL}

To evaluate the ability to transfer knowledge in FSCL, we evaluate TSC on miniImageNet, tieredImageNet and tieredImageNet\(\rightarrow\)CUB during incremental learning of ten 5W5S tasks. We first consider the backward transfer (BWT), which indicate the averaged influence that learning a new task has on the performance on the previous tasks. Then we examine the ability of the model to learn an extra 5W5S classification task consisting of novel class (NC, sampled from the query set) or base class (BC, sampled from the support set) in FSCL. As shown in Fig. \ref{FSCL_Improve_FSL}, TSC substantially improves the ability of FSL of the learned tasks, the query set and the support set from incremental tasks and examples in FSCL. Particularly, TSC can flexibly adapt to domain difference, since larger domain differences generally result in a more significant improvement of NC and BC. 


\begin{figure}[t]
	\centering
	\includegraphics[width=1\linewidth]{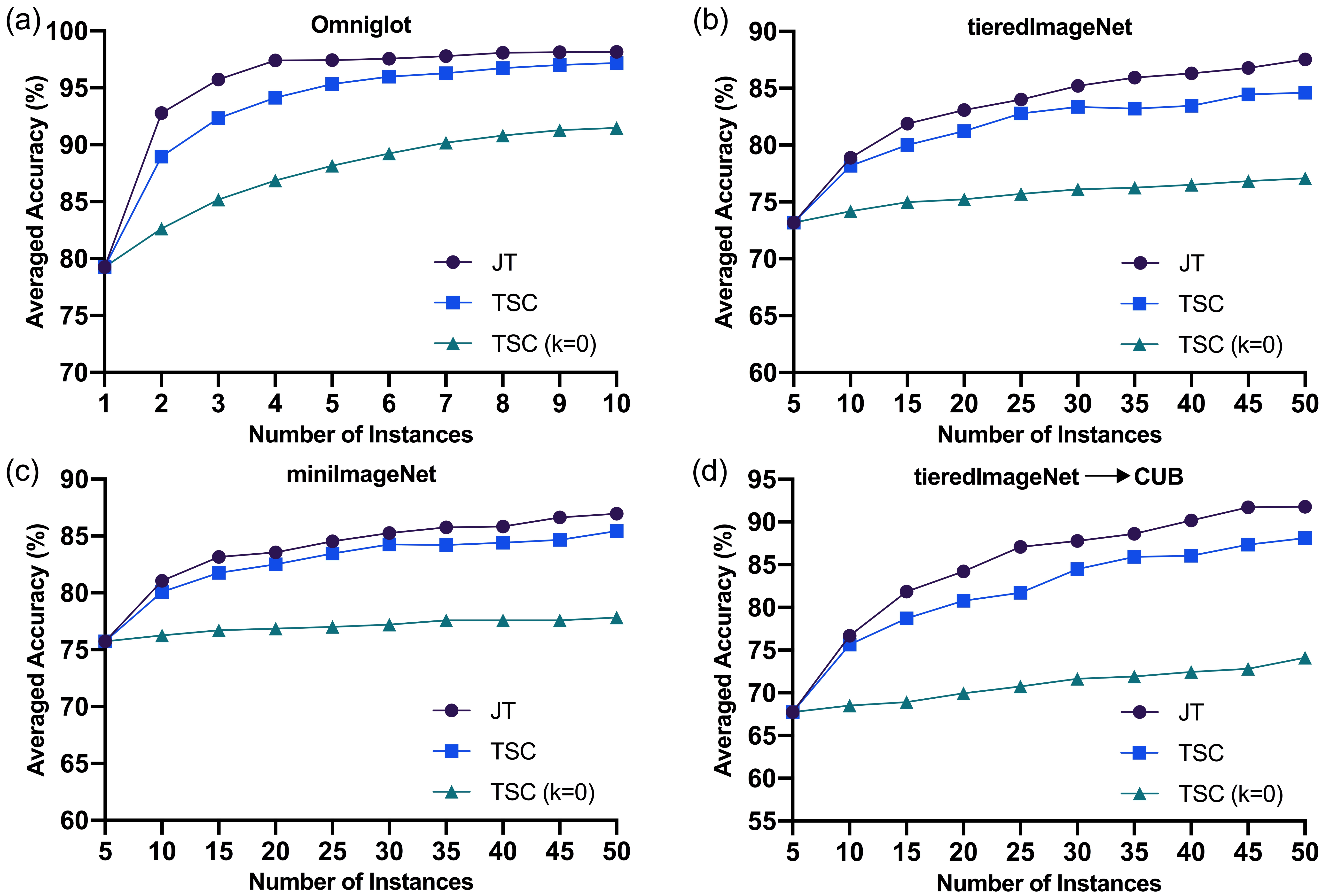}
	\caption{FSCL on new instance. FSL on one task is improved from sequentially seeing new instances by TSC.}
	\label{FSCL_NI}
   \vspace{-0.1cm}
\end{figure}

\subsection{FSCL of New Instance}
We now evaluate our framework on \textit{New Instance} \cite{lomonaco2017core50}, i.e. FSCL on incremental batches of a few new instances. Since all instances belong to the same task, the model sequentially learns new batch of instances \emph{without} replay of old batches. In Fig. \ref{FSCL_NI}, we randomly sample a task (20-class for Omniglot and 5-class for miniImageNet, tieredImageNet and CUB) and sequentially train the model with batches of new instances (1-instance / class for Omniglot, 5-instance / class for miniImageNet, tieredImageNet and CUB). After CL of several batches, we evaluate the performance of FSL on a new batch. The performance of TSC on one batch is continually improved from sequentially seeing new instances, only slightly lower than the joint training (JT) of all the instances ever seen. Note that TSC significantly outperforms TSC (\(k=0\)), which applies a fixed feature embedding layer and only updates the output layer. The results above demonstrate the effectiveness of TSC to continually learn new instances, incrementally improve FSL from CL, and adapt to domain differences.

\section{Conclusions}
In this work, we present a first systematic study to a challenging but realistic setting of few-shot continual learning (FSCL), in which the objectives of few-shot learning (FSL) and continual learning (CL) should be achieved simultaneously to improve each other. We present a solution to FSCL by drawing inspirations from the biological models of synaptic plasticity and replay. Extensive results on various datasets demonstrate the effectiveness of our approach for FSCL. Further work will focus on extending our method to other FSL approaches and CL scenarios.

{\small
\bibliographystyle{ieee_fullname}
\bibliography{egbib}
}

\clearpage
\appendix

\section{Few-Shot Continual Learning (FSCL) with Metric-Based Method}

In our setting of FSCL, a model aims to continually learn from a few examples, tackle a sequence of novel tasks, and incrementally improve the ability of few-shot learning (FSL). Although metric-based method is also a representative strategy for few-shot classification, continual learning (CL) of novel classes from a few examples is not easy. To continually improve few-shot generalization from incoming tasks and examples, metric-based FSL generally requires additional query data to update the feature embedding layer.

We notice that several recent works in this direction attempt to learn a few novel tasks / classes (sampled from a small query set) from their relations to the base tasks / classes (sampled from a large support set) and improve the joint prediction of both \cite{tao2020few,ren2019incremental}. We stress it is different from our setting of FSCL: The information of base tasks / classes and the support set might be unavailable in continual learning of novel tasks / classes, which is a practical assumption in real world scenarios. Also, prediction of only novel tasks / classes can better evaluate the ability of few-shot generalization in continual learning, compared with evaluation on the tasks / classes seen in the support set.

To achieve a considerable performance on a sequence of novel tasks / classes, another straightforward idea is to store all the training data and use a fixed feature embedding layer for joint prediction. However, such ``continual inference'' strategy cannot improve few-shot generalization from novel tasks and examples, and thus cannot adapt to domain differences or changing domains in continual learning.

\begin{figure}[ht]
	\centering
    \vspace{-0.1cm}
	\includegraphics[width=1\linewidth]{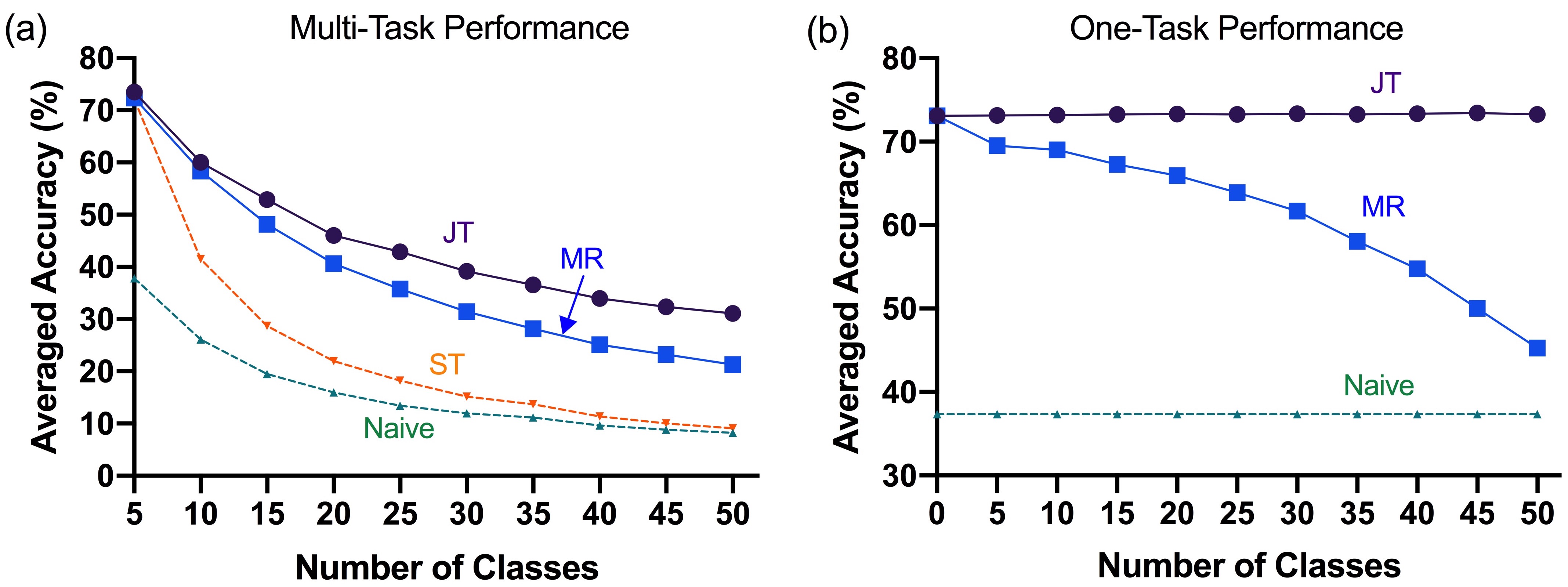}
	\vspace{-0.2cm}
	\caption{FSCL on miniImageNet with ProtoNet \textit{Linear Classifier}. (a) is the performance to infer all the incremental classes ever seen. (b) is the performance on one 5-way 5-shot task during continual learning in (a). JT: Joint Training; ST: Sequential Training; MR: ST with memory replay of all the data ever seen; Naive: A randomly-initialized network without pretraining (PT).}
	\label{fig1}
    \vspace{-0.1cm}
\end{figure}

To adapt metric-based FSL to FSCL, we propose a simple strategy: to use a linear classifier with softmax as the output layer in the FSCL stage, which has been shown as an effective strategy to improve generalization from \(S\) to \(D_T\), particularly in large domain differences \cite{chen2019closer}. Then we examine if the ability of FSL is interfered by CL in FSCL. We choose ProtoNet \cite{snell2017prototypical} as the base model of metric-based FSL and train such a classifier on \(S\). Next, we use a linear classifier with softmax as the output layer and continually train the classifier on a sequence of novel classification tasks. Similar to the observations in gradient-based FSL, the performance of FSCL with replay of all the old training data (ST+Replay) is significantly lower than joint training (JT) of all the tasks (Appendix Fig. 1, a). The ability of FSL also gradually decreases in CL (Appendix Fig. 1, b). 

The extension of ProtoNet \textit{Linear Classifier} can be easily implemented as the base model in TSC. We conduct the same FSCL experiments on \textit{New Class } as the main text. As shown in Appendix Table 1 and 2, TSC on ProtoNet \textit{Linear Classifier} achieves a better performance than joint training of all the tasks ever seen, and the ability of few-shot generalization is incrementally improved from continual learning.

\begin{table}[ht]
	\centering
	\caption{Averaged accuracy (\%) (top-1 / top-5) of class-incremental learning, averaged by 20 randomly sampled task sequences. }\smallskip
	\resizebox{1\linewidth}{!}{
	\begin{tabular}{cccccccc}
		\specialrule{0.01em}{1.2pt}{1.5pt}
   \multicolumn{1}{c}{}  &\multicolumn{2}{c}{miniImageNet*} & \multicolumn{2}{c}{tieredImageNet} &
		\multicolumn{2}{c}{tieredImageNet \(\rightarrow\) CUB}\\
		Methods & $ A_{30} $ & $ A_{50} $ & $ A_{30} $  & $ A_{50} $ & $ A_{30} $ & $ A_{50} $\\
		\specialrule{0.01em}{1.2pt}{1.5pt}
		Joint Training &36.65 / 69.01&27.86 / 56.03&37.95 / 70.93&29.25 / 60.47&34.79 / 68.72&27.53 / 57.75\\
        Memory Replay &27.36 / 54.69&17.05 / 36.81&26.78 / 53.71&16.08 / 35.85&21.96 / 48.76&14.47 / 33.53\\
        STC-M  &36.77 / 69.60&28.88 / 57.23&35.94 / 72.33&27.31 / 58.69&31.24 / 68.48&23.56 / 54.06\\
        TSC &\textbf{37.26} / \textbf{71.05}&\textbf{29.81} / \textbf{58.94}&37.39 / \textbf{73.24}&\textbf{30.33} / \textbf{62.77}&\textbf{35.80} / \textbf{69.49}&\textbf{27.79} / \textbf{58.04}\\
    
   \specialrule{0.01em}{1.2pt}{1.5pt}
	\end{tabular}
	}
	\label{table1}
\end{table}

\begin{table}[ht]
	\centering
        \vspace{-.2cm}
	\caption{Averaged absolute accuracy (\%)  / increased accuracy (\%)  of 5W5S task on various benchmark datasets in FSCL, averaged by more than 100 randomly sampled tasks.}\smallskip
	\resizebox{1\linewidth}{!}{
	\begin{tabular}{ccccc}
		\hline
		\multicolumn{2}{c}{}  & \multicolumn{1}{c}{Task 0} & \multicolumn{1}{c}{Task 5}  & \multicolumn{1}{c}{Task 10}  \\
		\specialrule{0.01em}{1.2pt}{1.5pt}
       \multirow{2}*{\tabincell{c}{miniImageNet*}}
        &\(k=20\) (Novel Class) &75.88  &77.92 / +2.04&78.47 / +2.59\\
        &\(k=100\) (Novel Class) &75.88  &78.85 / +2.97&\textbf{79.28 / +3.40}\\
		\specialrule{0.01em}{1.5pt}{1.5pt}
         \multirow{2}*{\tabincell{c}{tieredImageNet}}
        &\(k=20\) (Novel Class) &70.31  &73.01 / +2.70&73.51 / +3.20\\
        &\(k=100\) (Novel Class) &70.31  &73.83 / +3.52&\textbf{74.27 / +3.96}\\
		\specialrule{0.01em}{1.5pt}{1.5pt}
           \multirow{2}*{\tabincell{c}{tieredImageNet \\\(\rightarrow\) CUB}}
        &\(k=20\) (Novel Class) &60.77  &63.46 / +2.69&64.09 / +3.32\\
        &\(k=100\) (Novel Class) &60.77  &64.96 / +4.19&\textbf{65.71 / +4.94}\\
		\specialrule{0.01em}{1.5pt}{1.5pt}
	\end{tabular}
	}
	\label{table2}
	\vspace{-.2cm}
 \end{table}

\section{Implementations of Base Model}
Our implementation of MAML \cite{finn2017model} is validated in Appendix Table 3. miniImageNet is the original split proposed by \cite{vinyals2016matching}, which applies 64, 16 and 20 classes for training, validation and testing, respectively. To evaluate FSCL on large amounts of classes, we applies the entire 100 classes of miniImageNet (refer to as miniImageNet*) as the query set and the remained 900 classes in ImageNet as the support set. We follow similar implementations as \cite{chen2019closer,finn2017model} and reproduce the reported performance on various datasets and backbones \cite{finn2017model,chen2019closer, kim2019edge}. Note that we use 5-way classification for meta-training, because the ability of FSL might be improved from the incremental classes (ways) in FSCL. Then we apply the same implementation, but use ResNet-18 as the feature extractor, on miniImageNet* and tieredImageNet.

\begin{table}[ht]
	\centering
	\caption{Averaged 5W5S accuracy (\%) of MAML on various datasets. The performance of our results is averaged by more than 100 randomly sampled tasks.}\smallskip
	\resizebox{1.0\linewidth}{!}{
	\begin{tabular}{ccccc}
		\specialrule{0.01em}{1.5pt}{1.5pt}
         &Backbone &Omniglot&miniImageNet& tieredImageNet\\
        \specialrule{0.01em}{1.5pt}{2.0pt}
        MAML (Reported)& Conv-4  &\(99.9\) \cite{finn2017model} &\(63.15 \pm 0.91\) \cite{finn2017model} &\(70.30 \) \cite{kim2019edge}\\
        & & &\(62.71 \pm 0.71\) \cite{chen2019closer} & \\
       MAML (Ours)& Conv-4 & \(99.80 \pm 0.11\) &\(63.22 \pm 0.74\) & \(70.12 \pm 0.69\)\\
		\specialrule{0.01em}{2.0pt}{3.5pt}
	\end{tabular}
	}
    \resizebox{1.0\linewidth}{!}{
    	\begin{tabular}{ccccc}
		\specialrule{0.01em}{3.5pt}{1.5pt}
         &Backbone&miniImageNet&miniImageNet*& tieredImageNet\\
        \specialrule{0.01em}{1.5pt}{2.0pt}
        MAML (Reported)& ResNet-18& \(65.72 \pm 0.77\)  \cite{chen2019closer} &--&--\\
       MAML (Ours) & ResNet-18 &\(67.21 \pm 0.73\) & \(77.90 \pm 0.67\)& \(73.22\pm 0.60\)\\
		\specialrule{0.01em}{2.0pt}{1.5pt}
	\end{tabular}
	}
	\label{table3}
\end{table}

\section{Model Hyperparameters}
We use the Adam optimizer with \(beta1 = 0.5\) and \(beta 2 = 0.999\) to learn the classifier in all experiments. The hyperparameters to update fast /slow weights include the learning rate of \(\theta_{s}\) (\(\beta\)) and the iterations to learn \(\theta_f^{e}\) (\(k\)), both of which have been extensively evaluated in the main text. The model hyperparameters are summarized in Table \ref{Hyperparameters}.

\begin{table}[ht]
	\caption{Hyperparameters of our methods. }\smallskip
	\centering
	\resizebox{1\columnwidth}{!}{
		\smallskip\begin{tabular}{ccccc}
		\specialrule{0.01em}{1.2pt}{1.5pt}
        Hyperparameters & Omniglot & miniImageNet* & tieredImageNet & tieredImageNet \(\rightarrow\) CUB\\
	\specialrule{0.01em}{1.2pt}{1.5pt}
     Learning Rate & 0.0005  & 0.0005 &0.0005  & 0.0005 \\
     \(\beta\) &0.01  &0.01  &0.01  &0.01  \\
     \(k\) &100  &100 &100 &100   \\
     \(\lambda\) (STC-M) &0.1 &0.1 &0.1 &0.1  \\
     \(\lambda\) (TSC) &\(10^{-10}\) &\(10^{-10}\)  &\(10^{-10}\)  &\(10^{-10}\)   \\
     \(m\) &1 &1  &0.1  &0.1  \\
     \# of Epoch & 500& 500 & 500 & 500\\
	\specialrule{0.01em}{1.2pt}{1.5pt}
	\end{tabular}
}
	\label{Hyperparameters}
\end{table}

For the experiments of generative replay, we applies a 4-layer DCGAN architecture \cite{radford2015unsupervised} with WGAN-GP\cite{gulrajani2017improved} as the generative model to learn the old data distributions and replay generated data. However, a typical DCGAN architecture shows severe mode collapse to learn only a few training examples. This issue is caused by the BatchNorm layers in the generator and the discriminator, which can be replaced by InstanceNorm. The generator used in our implementation takes 3.80M parameters. The memory cost is equal to around 1853 images of the size \(64 \times 64\) (one image takes around 0.0082MB). To avoid the large memory cost of saving a generator, we sample 20 generated images after training on each class and store them in the memory buffer for replay.

\end{document}